\documentclass[sigconf]{acmart}
\AtBeginDocument{%
  }

\copyrightyear{2026}
\acmYear{2026}
\setcopyright{cc}
\setcctype{by}
\acmConference[DAC '26]{63rd ACM/IEEE Design Automation Conference}{July 26--29, 2026}{Long Beach, CA, USA}
\acmBooktitle{63rd ACM/IEEE Design Automation Conference (DAC '26), July 26--29, 2026, Long Beach, CA, USA}
\acmDOI{10.1145/3770743.3804433}
\acmISBN{979-8-4007-2254-7/2026/07}

\definecolor{olivegreen}{rgb}{0, 0.6, 0}

\usepackage{array, makecell} %
\usepackage{nicefrac}       %
\usepackage{booktabs}

\usepackage{multirow}
\usepackage{amsmath}
\usepackage{algorithm}
\usepackage{algorithmic}
\usepackage{subcaption}

\usepackage{pifont}%

\usepackage{comment}
\usepackage{xspace}         %
\usepackage{graphicx}%
\usepackage{lipsum}
\usepackage{multirow}
\usepackage{wrapfig}
\usepackage[capitalize,noabbrev]{cleveref}
\usepackage[table]{xcolor}

\newcommand{\aname}{TopGQ\xspace}

\newcommand{\ABSORB}{Dual-axis Scale Absorption\xspace}
\newcommand{\Absorb}{Dual-axis scale absorption\xspace}
\newcommand{\absorb}{dual-axis scale absorption\xspace}

\newcommand{\dq}{Degree-Quant\xspace}

\newcommand{\indegree}[1]{d(v_{#1})}
\newcommand{\neighbor}[1]{\mathcal{N}(v_{#1})}

\newcommand{\Iname}{TopPIN\xspace}
\newcommand{\Ifullname}{Topology-Aware Pairwise Index}
\newcommand{\Imath}{\mathrm{TopPIN}}

\newcommand{\XC}[1]{X_{\mathrm{c}}^{(#1)}}

\newcommand{\Xc}{X_{\mathrm{c}}}

\newcommand{\midscriptsize}{\fontsize{8.2pt}{9.2pt}\selectfont}
\newcommand{\qt}[1]{{\midscriptsize (#1)}}

\begin{document}

\title[TopGQ: Fast GNN Post-Training Quantization Leveraging Topology Information]{TopGQ: Fast GNN Post-Training Quantization Leveraging Topology Information}

\author{Dain Kwon}
\affiliation{%
  \institution{Seoul National University}
  \city{Seoul}
  \country{South Korea}
}
\email{dain.kwon@snu.ac.kr}

\author{Kanghyun Choi}
\affiliation{%
  \institution{Seoul National University}
  \city{Seoul}
  \country{South Korea}
}
\email{kanghyun.choi@snu.ac.kr}

\author{Hyeyoon Lee}
\affiliation{%
  \institution{Seoul National University}
  \city{Seoul}
  \country{South Korea}
}
\email{hylee817@snu.ac.kr}

\author{Sunjong Park}
\affiliation{%
  \institution{Seoul National University}
  \city{Seoul}
  \country{South Korea}
}
\email{ryan0507@snu.ac.kr}

\author{Seoyong Lee}
\affiliation{%
  \institution{Seoul National University}
  \city{Seoul}
  \country{South Korea}
}
\email{sylee2685@snu.ac.kr}

\author{Sukjin Kim}
\affiliation{%
  \institution{Seoul National University}
  \city{Seoul}
  \country{South Korea}
}
\email{iamksj1212@snu.ac.kr}

\author{Jinho Lee}
\orcid{0000-0003-4010-6611}
\affiliation{%
  \institution{Seoul National University}
  \city{Seoul}
  \country{South Korea}
}
\email{leejinho@snu.ac.kr}

\renewcommand{\shortauthors}{Dain Kwon et al.}

\begin{abstract}
Existing GNN quantization methods 
suffer from considerable quantization overhead, which severely limits their practical usage in real-world scenarios.
To this end, we present \textbf{\aname},
an accurate post-training GNN quantization framework, %
alleviating %
redundant quantization overhead. %
We propose \emph{\absorb}, which enables activation quantization along both the outer and inner dimensions by merging one into the adjacency matrix.
On top of that, we introduce \Iname, a proxy for nodes' local structure, and use it to group nodes with similar topology during quantization.
Experimental results %
show that \aname 
reduces quantization time by an order of magnitude while preserving accuracy.
The code is available at \url{https://github.com/meowrowan/TopGQ}.
\end{abstract}


\maketitle

\section{Introduction}

\begin{figure}[t]
    \includegraphics[width=0.42\textwidth]{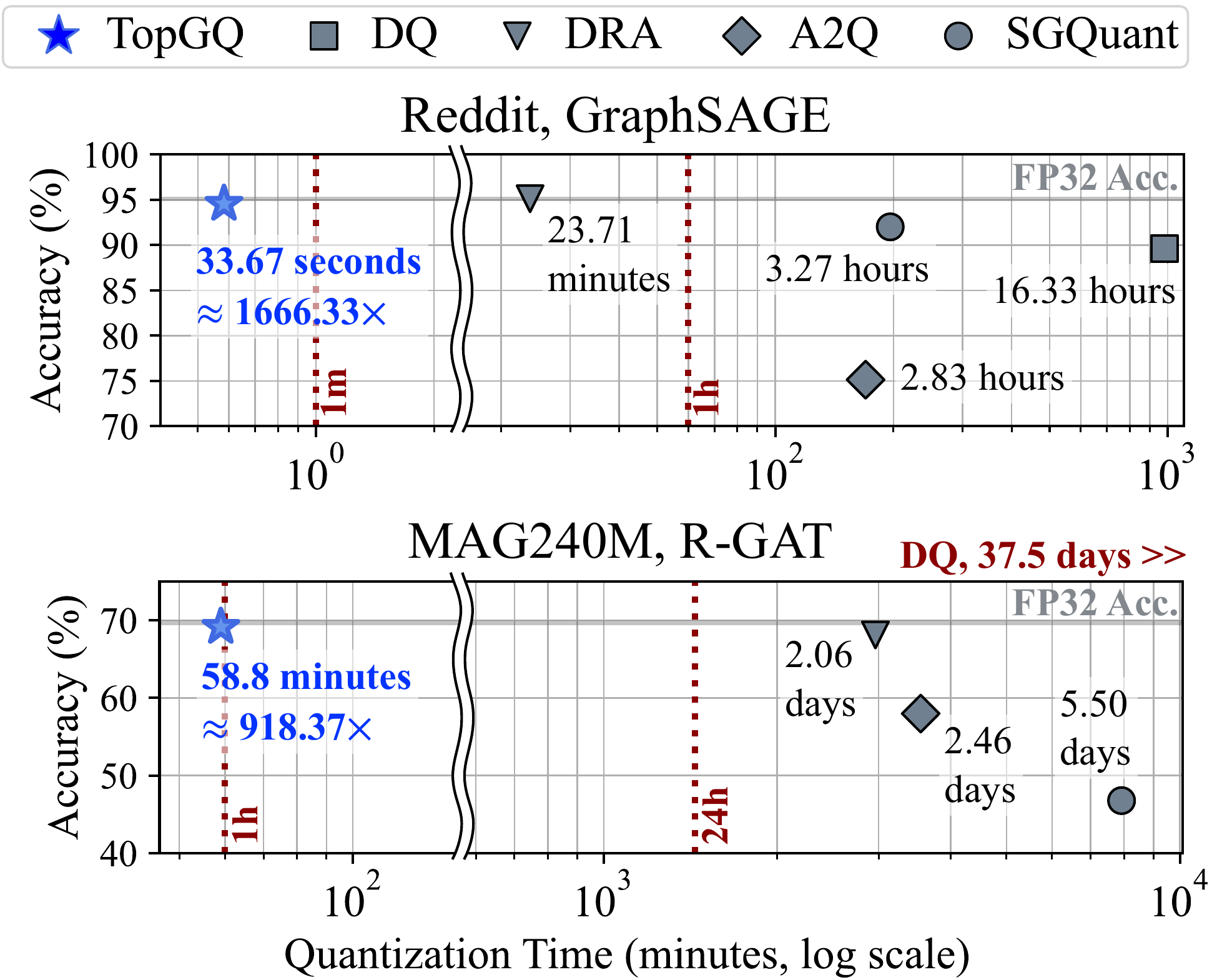}
    \caption{ Quantization time-accuracy trade-off plot with large-scale graph datasets.}
    \label{fig:intro_moti}
    \vspace{-5mm}
\end{figure}

Graph neural networks (GNNs) have attracted a great amount of attention due to their ability to process diverse unstructured data in diverse domains,
such as recommendation systems~\citep{zhang2023graph}, molecular interaction~\citep{nci1}, transportation networks~\citep{cao2020spectral}, and social network analysis~\citep{arazzi2023predicting}. 
Due to the rapid growth of the real-world graphs~\citep{liu2024scalable, ogb}, there is an urgent need to process them efficiently at scale.
One promising approach is quantization, which reduces memory and computational costs 
by using low-bit %
representations~\citep{ashkboos2024quarot, liu2024spinquant, lisvdquant, qdiffusion}. 

However, as illustrated in \cref{fig:intro_moti}, quantizing GNNs is difficult and time-consuming.
Existing work %
takes hours for medium-sized graphs such as Reddit, %
and days for larger graphs such as MAG240M, making it infeasible.
This overhead comes from handling outliers through extensive tuning or long training.
Quantization-aware training (QAT) methods involve such costly model retraining%
~\citep{sgquant, degreequant, epquant, a2q, qlr}.
While post-training quantization (PTQ) is known to be faster, existing methods~\citep{dra} still employ gradient-based iterations on quantization parameters, negating this advantage.

This large quantization time poses a major barrier to the deployment of GNN quantization 
in real-world scenarios, particularly when frequent model updates are required. 
Popular applications such as personalization and recommendation~\citep{inkstream, neutronstream, roland, reinc}
operate on large-scale graphs and benefit %
from quantization. 
However, these %
often require model updates on a minute-to-hour scale~\citep{monolith, ni2021prioritizing}, and excessive quantization time makes deployment impractical by canceling out the benefits of quantization.

For this, we present \aname, a PTQ method that achieves \emph{orders of magnitude faster} quantization with comparable or even better task performance. 
First, we show that node-wise quantization is preferable for GNNs (\cref{sec:challenges}), due to outlier nodes. 
However, node-wise quantization on the aggregation phase prevents the use of fast integer arithmetic.
Thus, we propose (1)
~\absorb, %
which enables fast and accurate integer matrix multiplication at aggregation %
by merging the scaling factors of each node into the adjacency matrix. %
We further propose (2)~\Iname, a fast-computable node index that captures its topology.
Using \Iname, we can rapidly assign quantization parameters to unseen nodes, ensuring fast inference.
Extensive experimental results validate that \aname outperforms current state-of-the-art baselines, 
achieving 
faster quantization 
while preserving accuracy and speed, establishing a new standard in GNN quantization.

\section{Background}
\label{sec:background}
\textbf{Graph neural networks.}
\label{sec:background:gnn}
Let $G=(V,E)$ be a directed graph with $n=|V|$ nodes, $v_1,\dots,v_n$. 
Denote $A \in \mathbb{R}^{n\times n}$ as the adjacency matrix, where $A_{ij} = 1_{(v_j,v_i) \in E}$. 
For node $v_i$, define its closed in-neighborhood as $\neighbor{i} = \{\,v_j \mid (v_j,v_i)\in E\}\cup\{v_i\}$, 
and let degree $\indegree{i}=|\neighbor{i}|$. We denote $D = \operatorname{diag}(\indegree{1}, \indegree{2}, \dots, \indegree{n})$ as the diagonal degree matrix,
$h_i$ as feature vector of $v_i$.

To embed topology, GNNs aggregate information from neighboring nodes $v_j \in \neighbor{i}$  
to update $h_i$. 
This procedure is referred to as the \emph{message-passing} algorithm, which consists of two steps: 
\emph{combination} and \emph{aggregation}. 
First, the hidden node feature $h_i^{(l)}$ is multiplied by the weight matrix $W^{(l)}$ in the $l$-th GNN layer. 
Next, the feature is aggregated to $h_i^{(l+1)}$ as follows:
\begin{align}
    h_i^{(l+1)} = \phi\Bigl(\bigoplus\nolimits_{\{j \mid v_j\in \neighbor{i} \}} 
        W^{(l)}\,h_j^{(l)}\Bigr),
\end{align}
where $\phi$ is an update function, and $\bigoplus$ is a permutation-invariant operator such as \textit{sum} or \textit{mean}.

The GNN computation can also be formulated in matrix form. 
Let $X^{(l)} = [\,h_1^{(l)}, \dots, h_{n}^{(l)}]^T \in \mathbb{R}^{n \times d_{l}}$ be the node feature matrix at layer $l$, 
and $W^{(l)}\in \mathbb{R}^{d_{l} \times d_{l+1}}$ be the weight matrix. 
Then, using %
the adjacency matrix $\tilde{A} \in \mathbb{R}^{n\times n}$, the combination and aggregation are:
\begin{align}
    \XC{l} =   X^{(l)}  W^{(l)},
    \quad
    X^{(l+1)} = \sigma\!\bigl(\tilde{A} \, \XC{l}\bigr), 
\end{align}
where $\sigma$ is a nonlinear function. 
The specific form of $\tilde{A}$ varies by GNN architecture. 
GCN~\citep{GCN} employs the normalized graph Laplacian $\tilde A = D^{-1/2}(A+I_n)D^{-1/2}$, 
while GIN~\citep{gin} uses the binary matrix $\tilde A =A + I_n$. 
GraphSAGE~\citep{graphsage} differs by sampling a subset of neighbors instead of using the entire neighborhood at aggregation.

\textbf{Transductive and inductive settings.}
GNN training operates in either a transductive or an inductive setting. 
In the transductive setting, the full graph (e.g., features and topology of test nodes) is available during training, except for the test node labels.
Thus, inference can be done with precomputed embeddings~\citep{oblivgnn}, leaving little room for acceleration %
benefits. 
In contrast, the inductive setting introduces unseen nodes or graphs at test time, requiring computation for node embeddings during inference.
Consequently, GNN quantization is especially valuable %
in inductive settings, where reducing computation and memory directly accelerates inference.
Moreover, the inductive setting better reflects practical real-world scenarios where graphs evolve or differ from those used for training, such as social networks and recommendation systems with new users, %
or molecular property prediction for unseen molecules.

\label{subsec:quant}
\textbf{Quantization} replaces high-precision floating-point operations with low-bit integer operations, reducing computational cost and memory usage. 
We adopt uniform quantization with scale ($s$) and zero-point ($z$). 
Given a tensor $X$, element $x \in X$ is quantized as:
\begin{align}
x^{q} &= 
           \mathrm{clamp}\Big(\Big\lfloor \frac{1}{s} \cdot (x - z) \Bigr\rceil,\, q_{\min},\, q_{\max}\Big), \quad
    s = \frac{x_{\max} - x_{\min}}{q_{\max}-q_{\min}},   %
\end{align}
$q_{\min}$ and $q_{\max}$ are the minimum and maximum integer values in $k$-bit representation, and $\lfloor \cdot \rceil$ denotes rounding. 
Quantization operates at various granularities, such as per-tensor, per-row, or per-column.
Finer granularity can reduce quantization error by constraining the effect of outliers to only a subset of values.
However, we cannot freely choose per-row or per-column granularity, %
since the scale factors must follow the outer dimension of the GEMM for dequantization.
This induces challenges for GNN quantization (\Cref{sec:challenges}).

Quantization can also be categorized into post-training quantization (PTQ) and quantization-aware training (QAT),
depending on whether quantization is %
applied after training or incorporated during training.
QAT iteratively updates the model weights using backpropagated gradients under simulated quantization, whereas PTQ calibrates scale and zero-point without modifying the pretrained weights, making it substantially faster in practice.
As our method is PTQ-based, it %
inherits benefits from this efficiency.

\section{Motivations \& Challenges} %

\label{sec:challenges}
\begin{figure}[b]
    \centering
    \begin{subfigure}[b]{0.45\columnwidth}
        \centering
        \includegraphics[width=\linewidth]{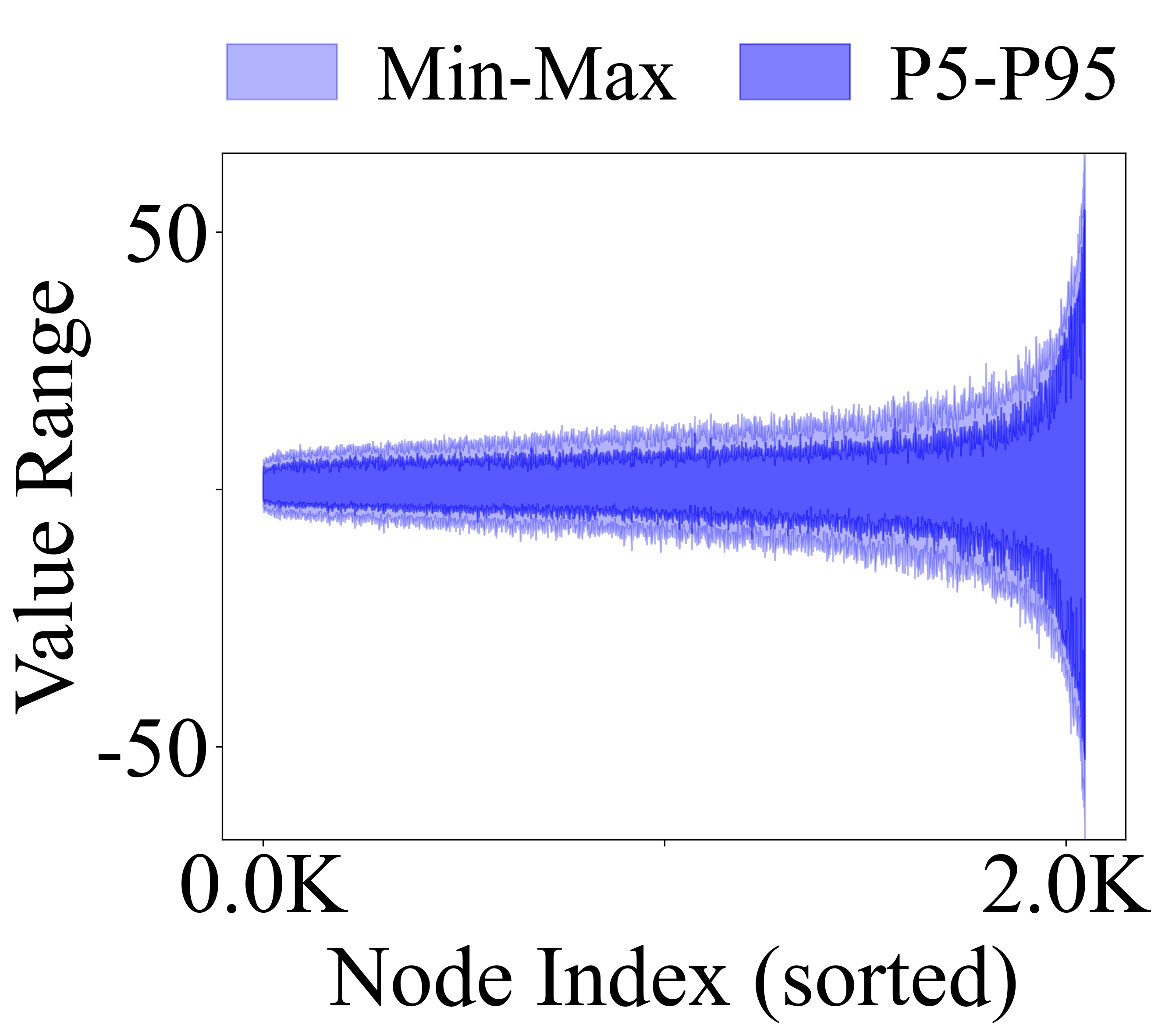}
        \caption{Reddit, GCN}
        \label{fig:nodewise-reddit}
    \end{subfigure}
    \begin{subfigure}[b]{0.45\columnwidth}
        \centering
        \includegraphics[width=\linewidth]{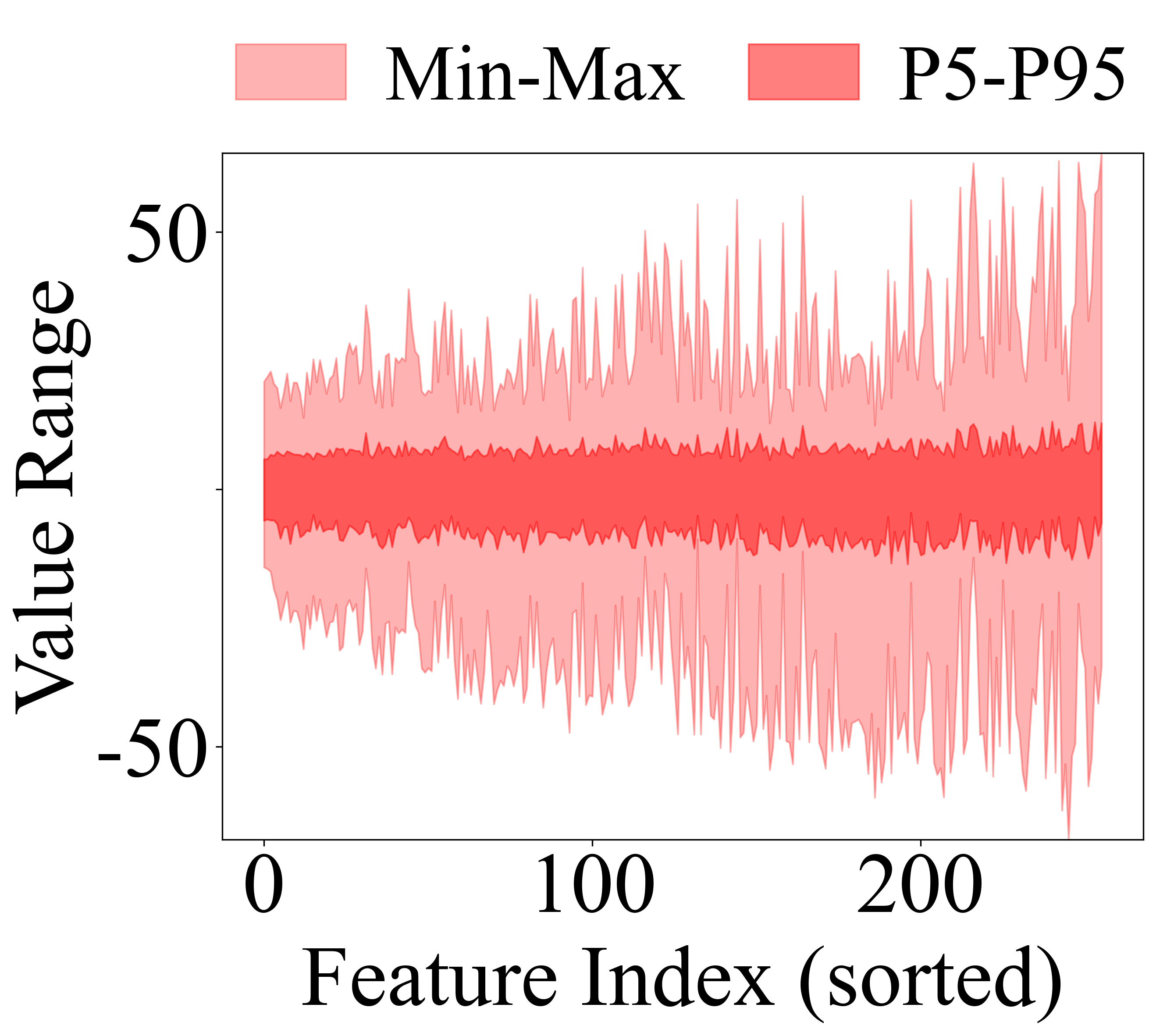}
        \caption{Reddit, GCN}
        \label{fig:featurewise-reddit}
    \end{subfigure}
    \begin{subfigure}[b]{0.45\columnwidth}
        \centering
        \includegraphics[width=\linewidth]{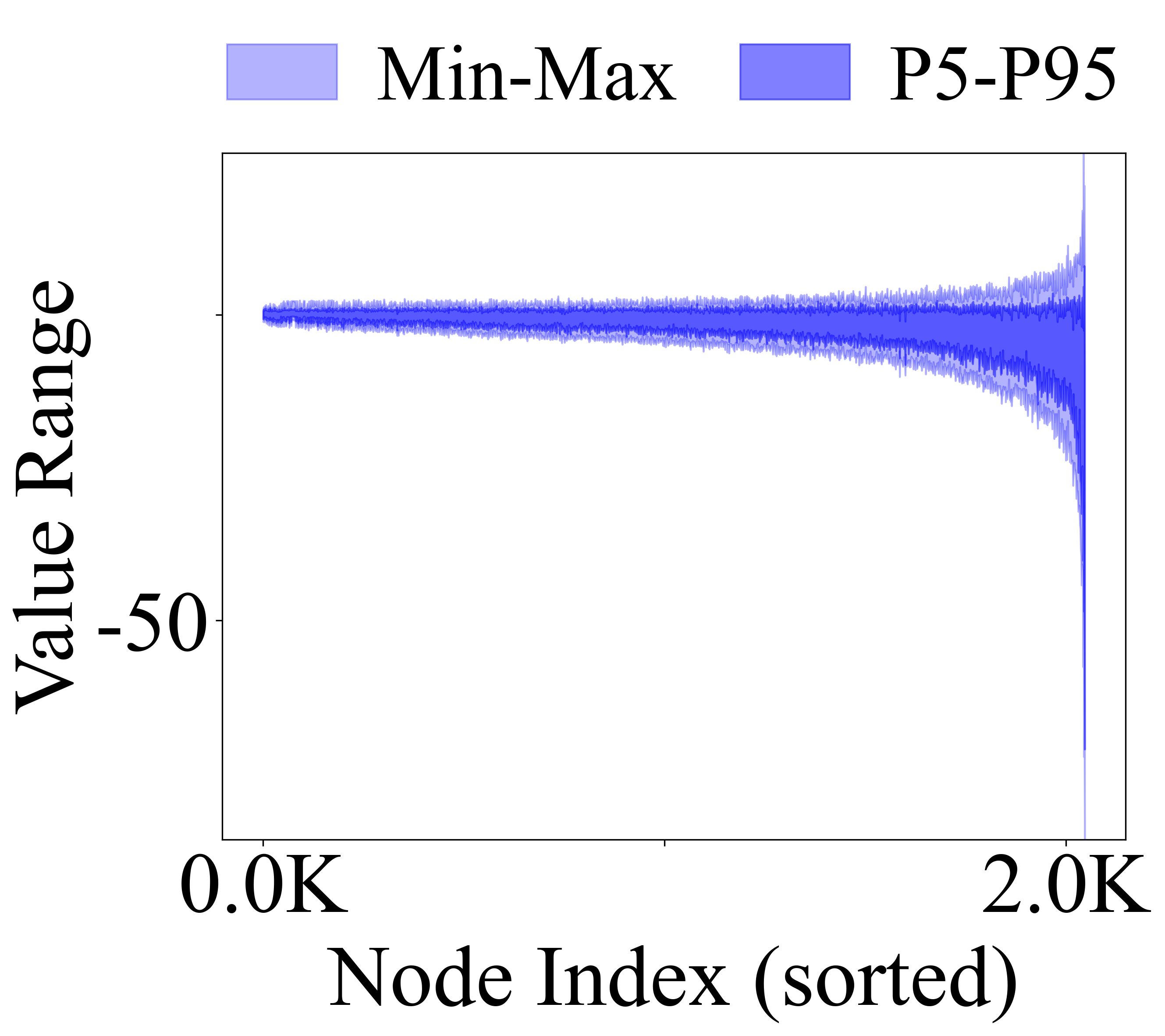}
        \caption{ogbn-products, GraphSAGE}
        \label{fig:nodewise-ogbn}
    \end{subfigure}
    \begin{subfigure}[b]{0.45\columnwidth}
        \centering
        \includegraphics[width=\linewidth]{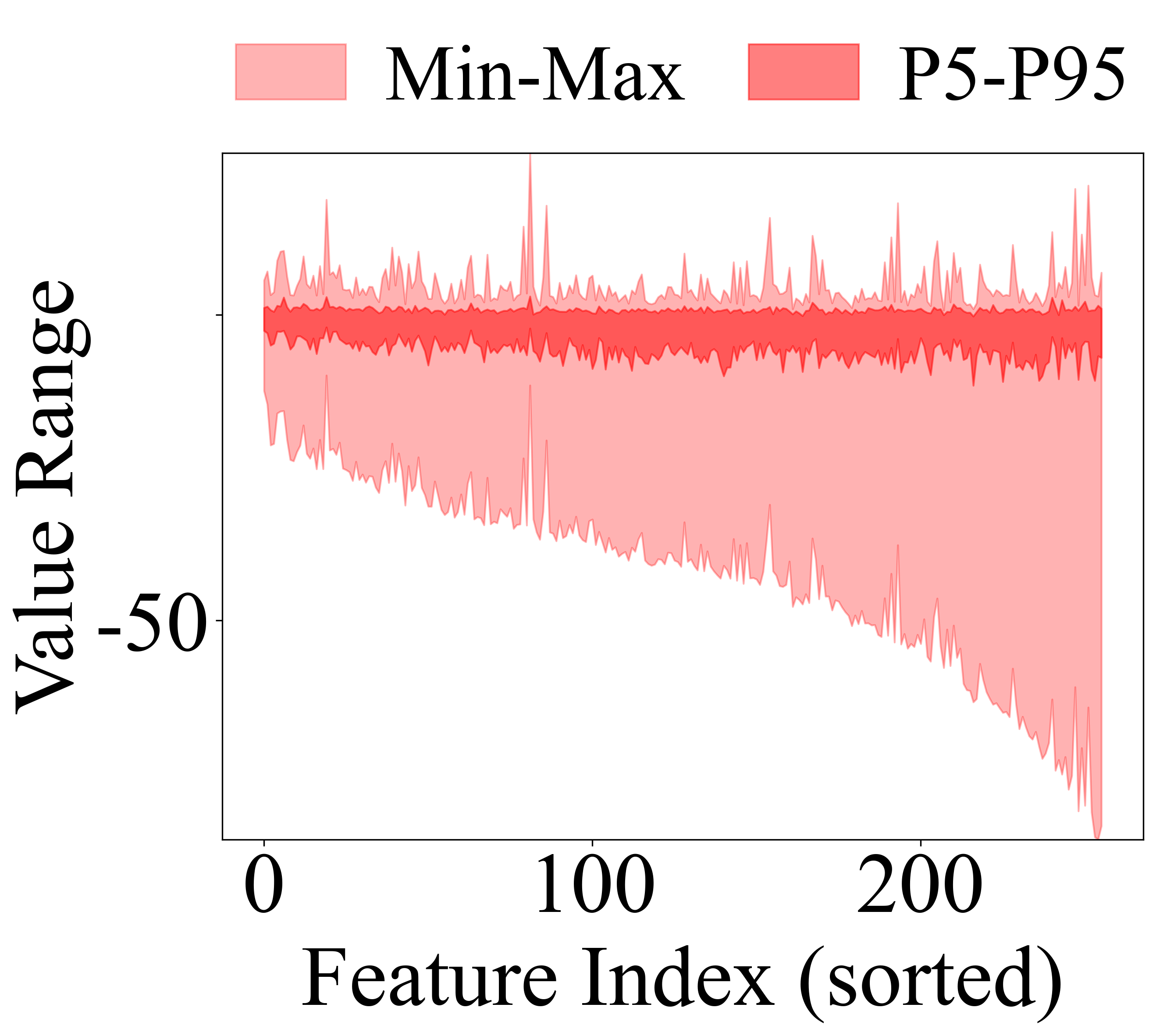}
        \caption{ogbn-products, GraphSAGE}
        \label{fig:featurewise-ogbn}
    \end{subfigure}

    \caption{Node-wise(blue, left) and feature-wise(red, right) range plot, sorted in ascending order. `Node Index' indicates each node, and `Feature Index' indicates each feature dimension. Each plot shows the min-max range and the 5th-95th percentile range of the values within the same dimension.}
    \label{fig:rowwise}
    \vspace{-5mm}
\end{figure}
GNN quantization requires special consideration due to its unique message-passing mechanism.
In particular, the accumulation of neighborhood information induces diversity across nodes, making node-wise quantization a preferred approach.
\cref{fig:rowwise} illustrates such behavior by comparing the activation range within each node dimension (\cref{fig:nodewise-reddit,fig:nodewise-ogbn}) and feature dimension (\cref{fig:featurewise-reddit,fig:featurewise-ogbn}).
\cref{fig:nodewise-reddit,fig:nodewise-ogbn} show that node-wise ranges are more concentrated,
with %
the 5th–95th percentile range close to the min–max range.
This indicates that each %
group exhibits a suitable range for quantization, as there are no outliers far exceeding the majority distribution.
However, in the feature-wise plots (\cref{fig:featurewise-reddit,fig:featurewise-ogbn}), each min-max range is much broader, while 95\% of the values exist within a narrower interval.
This distribution is more prone to outliers far exceeding the majority range of values, leading to wasted quantization bins and higher error.
This makes node-wise quantization a better choice for GNN activations.

Based on the observation, we assign different quantization scales to the group of nodes for %
$X$ in both the combination and the aggregation phase. %
Enabling such a method in the {\em combination} phase is relatively straightforward. 
In fact, existing methods~\citep{a2q, sgquant} already employ node-wise quantization for combination ($X \cdot W$):
\begin{equation}
\resizebox{0.92\hsize}{!}{$
\displaystyle
    X \cdot W 
     \approx 
    \operatorname{diag}(S_X) \cdot X^Q \cdot W^Q \cdot \operatorname{diag}(S_W)
    = (S_X \cdot S_W^\top) \odot (X^Q \cdot W^Q),
    \label{eq:absorbnotneeded}
$}
\end{equation}
where $S_X \in \mathbb{R}^{n \times 1}$ is the node-wise scale of $X$, $S_W \in \mathbb{R}^{d \times 1}$ the feature-wise scale of $W$, and $\odot$ denotes the element-wise (Hadamard) product. 
As $X$ is quantized node-wise and $W$ feature-wise, their multiplication remains 
a standard integer GEMM,
with the scales applied afterwards for dequantization, 
maintaining high throughput.

\paragraph{Challenge 1: Quantization along inner dimensions.} 
By contrast, node-wise quantization in the {\em aggregation} phase is challenging. 
Applying node-wise quantization for the aggregation step ($ \tilde{A} \cdot \Xc$),
\begin{align}
    \tilde{A} \cdot \Xc \;\approx\; 
    \operatorname{diag}(S_{\tilde{A}}) \cdot \tilde{A}^Q \;\cdot\; 
    \operatorname{diag}(S_{\Xc}) \cdot \Xc^Q,
    \label{eq:absorbneeded}
\end{align}
introduces the diagonal matrix $\operatorname{diag}(S_{\Xc})$ within the multiplication.
Unlike Equation \ref{eq:absorbnotneeded}, this cannot be computed using integer matrix multiplication units%
~\citep{jacob2018quantization}. %
Thus, existing GNN quantization~\citep{a2q, sgquant} methods simply choose to employ column-wise quantization to $\Xc$.
While this ensures acceleration, it may fail to preserve the precision of activations. %
To achieve the precision benefits of node-wise quantization while maintaining the efficiency of integer GEMM, \aname proposes a novel method, 
\emph{\absorb} (\cref{sec:main:absorb}).

\paragraph{Challenge 2: Generalization on unseen nodes.} 
For practical inductive settings (\cref{sec:background}), %
GNN encounters unseen nodes at inference. 
There are two ways to get quantization parameters for such nodes:

\emph{(i) On-the-Fly Quantization Parameter Computation.}
A straightforward approach is to dynamically compute quantization parameters per node during inference. 
For each activation, every row of $X^{(l)}$ and $\XC{l}$ is scanned, and the minimum and maximum values of each node are empirically determined to obtain scales and zero-points. 
While this ensures low quantization error, %
it is less preferred as the additional runtime overhead can offset the efficiency gains. %

\emph{(ii) Precomputed Mapping.}
An alternative is to precompute a set of quantization parameters from train nodes at calibration time, and map each unseen node to one of them at inference.
Before inference, %
a simple lookup can retrieve and prepare appropriate parameters for each activation. 
Nonetheless, this requires an accurate low-complexity \emph{node index} $\phi(\cdot)$ such that nodes with similar index values exhibit similar feature statistics. 
\aname chooses this %
mapping approach, 
where we design a novel \emph{\Ifullname~(\Iname)} that 
uses local topology for lightweight computation (\cref{sec:dain}). 
\Iname ensures that unseen nodes are assigned adequate quantization parameters at low inference overhead.

\section{\aname Methodology}
\label{sec:main}

\subsection{Selective \ABSORB}
\label{sec:main:absorb}

\begin{figure}[b]
    \centering
    \includegraphics[width=0.405\textwidth]{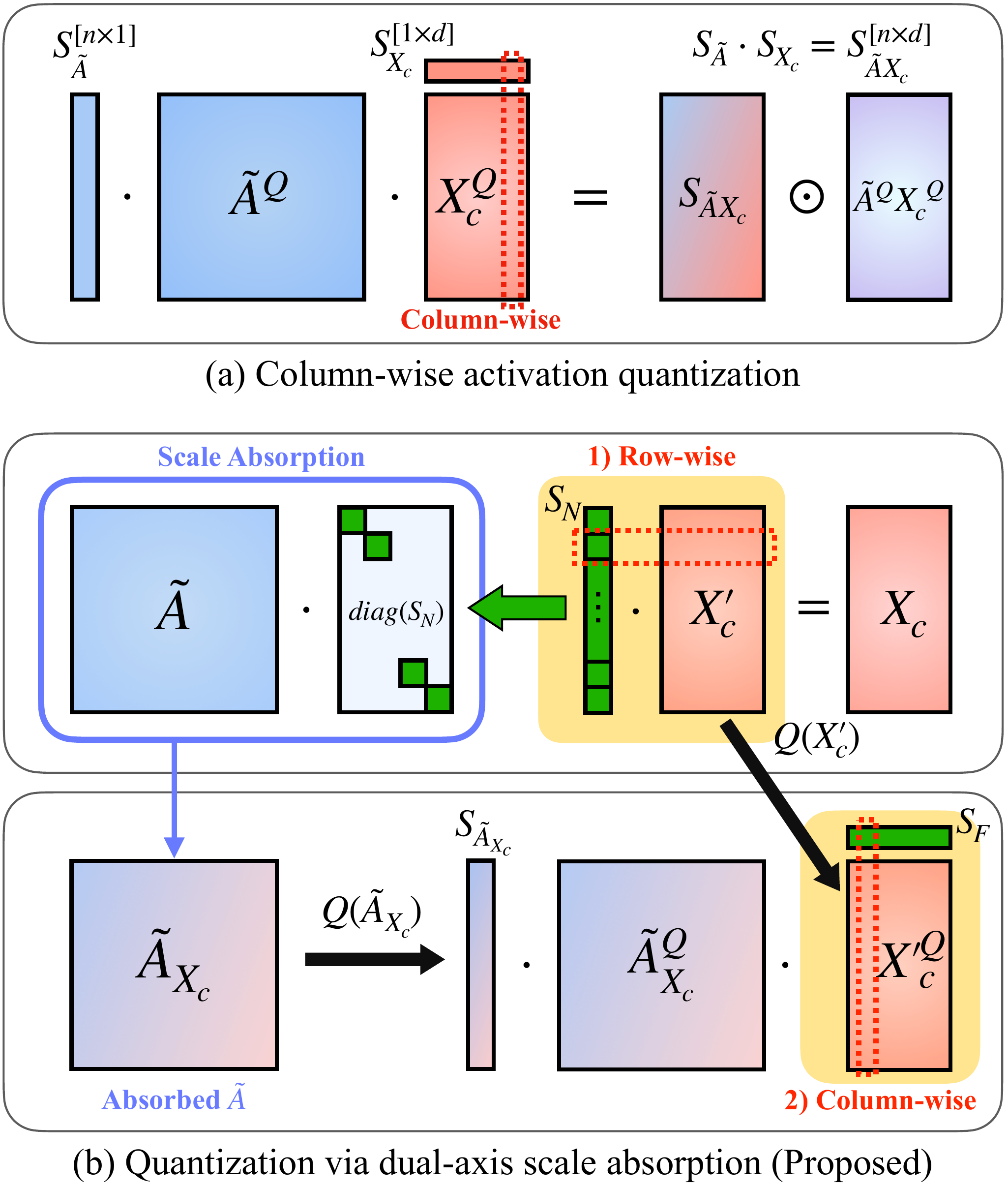}
    \caption{Comparing quantization at \emph{aggregation} phase.
    }
    \label{fig:dual_axis}
\end{figure}

\begin{figure*}[t]
    \centering
    \includegraphics[width=\textwidth]{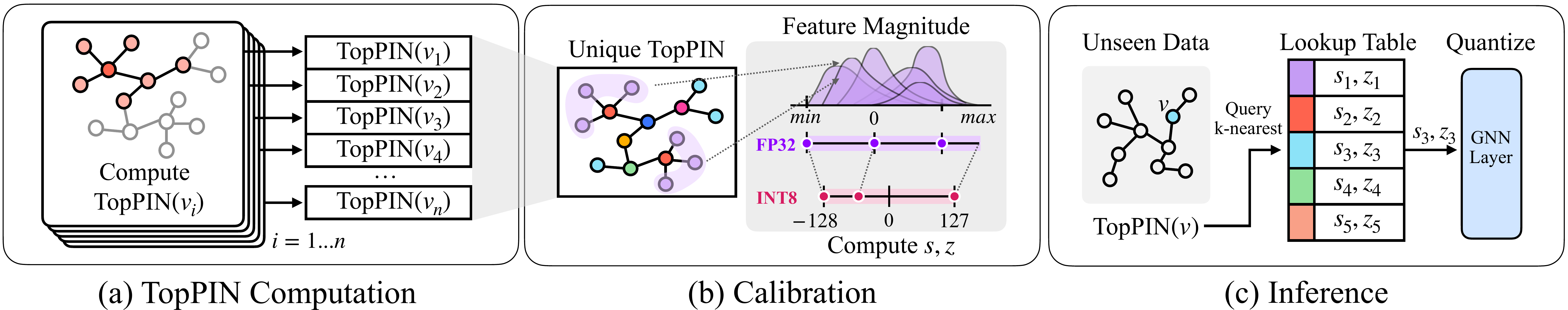}
\caption{ Quantization process with \Iname. (a) shows node group generation through the proposed topological proxy, \Iname.
Each color is used to denote each group. %
(b) shows the calibration process to obtain quantization parameters for each group.  %
(c) demonstrates how inference is done on unseen data, using the quantization parameters of the nearest groups with interpolation.} 
\vspace{-3mm}
\label{fig:overall} 
\end{figure*}

To account for the differing magnitude of node features (\cref{fig:rowwise}), we employ a node-wise scale factor $S_N \in \mathbb{R}^{N \times 1}$, where $S_N$ consists of the maximum feature value for each node. 
Specifically, we scale $\Xc$ to $\Xc'$ with $S_N$, i.e., $\Xc' = \operatorname{diag}^{-1}(S_N) \cdot \Xc $.
Then, to eliminate any terms preventing integer operations, $S_N$ is merged to the given static adjacency matrix, $\tilde{A}\in \mathbb{R}^{N \times N}$.
The operation is as follows:
\begin{align}
    \tilde{A} \cdot \Xc 
    &= (\tilde{A} \cdot \operatorname{diag}(S_N)) \cdot \Xc'
    = \tilde{A}_{\Xc} \cdot \Xc'.
    \label{eq:row-wise-scaling}
\end{align}

After merging $S_N$ to $\tilde{A}$, we can conduct integer matrix multiplication for two matrices, 
$\tilde{A}_{\Xc}$ and $\Xc'$ with corresponding quantization parameters $S_{\tilde{A}_{\Xc}} \in \mathbb{R}^{N \times 1}$, and $S_{\Xc'} \in \mathbb{R}^{1 \times d}$: %
\begin{align}
    \tilde{A}_{\Xc} \cdot \Xc' &\approx (\operatorname{diag}(S_{\tilde{A}_{\Xc}}) \cdot \tilde{A}_{\Xc}^Q) \cdot (\Xc'^Q \cdot diag(S_{\Xc'})) \\
    &=(S_{\tilde{A}_{\Xc}} \cdot S_{\Xc'})\odot(\tilde{A}_{\Xc}^Q \cdot \Xc'^Q).
    \label{eq:absorbquant}
\end{align}

In the calibration process, \aname adaptively chooses between dual-axis and feature-wise quantization for $\Xc$ for each GNN layer. 
\aname evaluates both configurations by measuring the mean squared error (MSE) between the original floating-point activations and their quantized counterparts. 
The configuration with lower MSE is saved for inference. 
When \absorb is selected, the scaling elements for $S_N$ are calibrated like the quantization parameters.
When \absorb is used at inference, $\Xc$ can be immediately quantized with $S_N \cdot S_{\Xc'} \in \mathbb{R}^{N \times d}$, which acts like an element-wise quantization parameters for $\Xc$. 

\Absorb mimics the effect of node-wise quantization, while actually using feature-wise quantization to be compatible with integer matrix multiplication.
This design allows 
advantages of node-wise scaling without sacrificing inference efficiency.

\subsection{\Iname: A Fast Index for Unseen Nodes}
\label{sec:dain}

To support quantization in inductive settings,
we devise \Iname, a lightweight index that maps unseen nodes to existing train set nodes used at calibration. 
The formulation of \Iname is as follows:
\begin{align}
    \Imath(v) %
                      &= \Bigl(\;\indegree{},\; \frac{1}{\indegree{}} \sum_{v_k \in \neighbor{}} \frac{1}{\indegree{k}} \;\Bigr).
\end{align}

We observe that GNN aggregation creates distinct activation patterns for different nodes based on their local topology. 
When a GNN repeatedly aggregates neighborhood information through layers, the variance of each node's activation accumulates differently depending on its position in the graph.
Specifically, we can characterize this accumulation through a node index function $\phi: V \to \mathbb{R}$.
For unnormalized GNNs ($\tilde{A}=A+I_n$), this accumulation is proportional to the number of paths reaching the nodes: 
\begin{align}
\phi(v)=\Sigma_{v_{k_1} \in N(v)}\Sigma_{v_{k_2} \in N(v_{k_1})} \quad ... \quad \Sigma_{v_{k_l} \in N(v_{k_{l-1}})}1, 
\end{align}
where nodes with denser neighborhoods accumulate more variance.

This idea suggests that nodes with similar $\phi$ values will exhibit similar activation distributions after multiple layers of aggregation.
Rather than using uniform quantization parameters across all nodes, we can group nodes by their expected activation patterns based on local graph structure. 
This insight motivates our \Iname index, which captures these topological patterns to enable efficient node-wise quantization by assigning similar quantization parameters to nodes that share similar neighborhood aggregation characteristics.
To derive a practical approximation of $\phi(v)$ that forms our \Iname index,
let $d(v)$ denote the indegree of node $v$, and we approximate all summations beyond the first term as a constant $C_1$:
\begin{equation}
\phi(v) \approx \sum_{v_{k_1} \in \mathcal{N}(v)} C_1 = d(v) \cdot C_1.
\end{equation}
For normalized GNNs ($\tilde{A}=D^{-1/2}(A+I_n)D^{-1/2} $), the only difference is that degree normalization factors propagate through the aggregation.
In this case, 
we approximate the summand of the second summation as a constant $C_2$:
\begin{equation}
\resizebox{0.98\hsize}{!}{$
\displaystyle
\phi(v) \approx \frac{1}{d(v)} \sum_{v_{k_1} \in \neighbor{}} 
\biggl( \;
\frac{1}{d(v_{k_1})^2} 
\sum_{v_{k_2} \in \neighbor{k_1}} C_2
\biggr)
=\frac{1}{d(v)}\sum_{v_{k_1} \in \neighbor{}} \frac{C_2}{d({v_{k_1}})}
$}
\end{equation}

These approximations lead to our final \Iname formulation, which effectively balances accuracy and efficiency. 
$\Imath(v)$ %
can be applied to various GNN architectures. %
Our empirical observations confirm that these first-order approximations capture most of the benefits with minimal computational overhead, making \Iname practical for large-scale inductive GNN quantization.

\begin{table*}[t]
  \centering
  \captionof{table}{Comparison of node classification task with graphs of varying sizes}
  \label{tab:large_graph_node}
  \setlength{\tabcolsep}{3.5pt}
  \resizebox{.98\textwidth}{!}{%
  \begin{tabular}{c c c *{16}{c}}
    \toprule
    \multirow{3}{*}{Bit}
      & \multirow{3}{*}{Method}
      & \multirow{3}{*}{Type}
      & \multicolumn{4}{c}{Cora}
      & \multicolumn{4}{c}{Citeseer}
      & \multicolumn{4}{c}{Reddit}
      & \multicolumn{4}{c}{ogbn-products} \\
    \cmidrule(lr){4-7}\cmidrule(lr){8-11}\cmidrule(lr){12-15}\cmidrule(lr){16-19}
      &
      \multicolumn{1}{c}{}
      && \multicolumn{2}{c}{GCN} & \multicolumn{2}{c}{GraphSAGE}
      & \multicolumn{2}{c}{GCN} & \multicolumn{2}{c}{GraphSAGE}
      & \multicolumn{2}{c}{GCN} & \multicolumn{2}{c}{GraphSAGE}
      & \multicolumn{2}{c}{GCN} & \multicolumn{2}{c}{GraphSAGE} \\
    \cmidrule(lr){4-5}\cmidrule(lr){6-7}\cmidrule(lr){8-9}\cmidrule(lr){10-11}
    \cmidrule(lr){12-13}\cmidrule(lr){14-15}\cmidrule(lr){16-17}\cmidrule(lr){18-19}
      &&
      & Acc. & Q.Time & Acc. & Q.Time
      & Acc. & Q.Time & Acc. & Q.Time
      & Acc. & Q.Time & Acc. & Q.Time
      & Acc. & Q.Time & Acc. & Q.Time \\
    \midrule
    \rowcolor{gray!10}\cellcolor{white}
      & FP32  & --
      & 80.14 & --   & 77.02 & --
      & 76.46 & --   & 76.34 & --
      & 94.40 & --   & 95.09 & --
      & 71.25 & --   & 70.33 & -- \\
    \midrule

    \multirow{5}{*}{INT8}
    & SGQ & QAT
        & 79.93 & \qt{6.4\textbf{s}}     & 76.28 & \qt{8.24\textbf{s}}
        & 76.21 & \qt{9.3\textbf{s}}     & 76.06 & \qt{14.84\textbf{s}}
        & 92.10 & \qt{4.64\textbf{m}}    & 92.01 & \qt{3.27\textbf{h}}
        & 39.13 & \qt{2.52\textbf{m}}    & 58.80 & \qt{2.11\textbf{h}} \\
      & DQ & QAT
        & 78.94 & \qt{11.5\textbf{s}}    & 75.50 & \qt{23.87\textbf{s}}
        & 75.62 & \qt{25.7\textbf{s}}    & 74.77 & \qt{69.95\textbf{s}}
        & 87.01 & \qt{10.59\textbf{m}}   & 90.53 & \qt{16.35\textbf{h}}
        & \textbf{72.34} & \qt{14.05\textbf{m}} & 70.17 & \qt{6.55\textbf{h}} \\
      & $A^2Q$ &  QAT
        & 79.66 & \qt{3.5\textbf{s}}     & \textbf{76.94} & \qt{4.56\textbf{s}}
        & 75.72 & \qt{3.5\textbf{s}}     & 75.08 & \qt{4.96\textbf{s}}
        & 73.71 & \qt{4.12\textbf{m}}    & 75.13 & \qt{2.83\textbf{h}}
        & 50.78 & \qt{83.94\textbf{s}}   & 60.15 & \qt{1.67\textbf{h}} \\
      & DRA & PTQ
        & 79.76 & \qt{1.7\textbf{s}}     & 76.46 & \qt{3.11\textbf{s}}
        & 76.26 & \qt{1.6\textbf{s}}     & 75.74 & \qt{3.00\textbf{s}}
        & 93.15 & \qt{42.99\textbf{s}}   & 94.36 & \qt{23.71\textbf{m}}
        & 36.22 & \qt{41.63\textbf{s}}   & 47.70 & \qt{44.97\textbf{m}} \\
    \rowcolor{blue!10}\cellcolor{white}
      & \aname & PTQ
        & \textbf{79.96} & \qt{0.2\textbf{s}}   & 76.86 & \qt{0.54\textbf{s}}
        & \textbf{76.52} & \qt{0.2\textbf{s}}   & \textbf{76.32} & \qt{0.56\textbf{s}}
        & \textbf{94.41} & \qt{1.88\textbf{\textcolor{red}{s}}}
        & \textbf{94.55}          & \qt{35.79\textbf{\textcolor{red}{s}}}
        & 71.33          & \qt{1.16\textbf{\textcolor{red}{s}}}
        & \textbf{70.31} & \qt{34.88\textbf{\textcolor{red}{s}}} \\
    \midrule

    \multirow{5}{*}{INT4}
    & SGQ & QAT
        & 78.73 & \qt{6.4\textbf{s}}     & 75.52 & \qt{8.41\textbf{s}}
        & \textbf{76.31} & \qt{10.4\textbf{s}}    & \textbf{75.94} & \qt{14.65\textbf{s}}
        & 43.00 & \qt{4.84\textbf{m}}    & 87.42 & \qt{3.27\textbf{h}}
        &  6.14 & \qt{2.57\textbf{m}}    & 27.95 & \qt{2.13\textbf{h}} \\
      & DQ & QAT
        & 78.54 & \qt{11.5\textbf{s}}    & 74.36 & \qt{23.49\textbf{s}}
        & 23.54 & \qt{25.5\textbf{s}}    & 74.99 & \qt{69.91\textbf{s}}
        & 64.18 & \qt{10.55\textbf{m}}   & 89.61 & \qt{16.33\textbf{h}}
        & 36.66 & \qt{13.93\textbf{m}}   & \textbf{69.90} & \qt{6.52\textbf{h}} \\
      & $A^2Q$ & QAT
        & 50.00 & \qt{3.6\textbf{s}}     & 74.66 & \qt{4.65\textbf{s}}
        & 43.52 & \qt{3.5\textbf{s}}     & 73.00 & \qt{5.01\textbf{s}}
        & 23.24 & \qt{4.12\textbf{m}}    & 67.94 & \qt{2.83\textbf{h}}
        & 25.95 & \qt{83.30\textbf{s}}   & 31.32 & \qt{1.66\textbf{h}} \\
      & DRA & PTQ
        & 77.02 & \qt{1.7\textbf{s}}     & 76.18 & \qt{3.24\textbf{s}}
        & 74.10 & \qt{1.6\textbf{s}}     & 74.60 & \qt{2.93\textbf{s}}
        &  1.75 & \qt{42.82\textbf{s}}   &  5.31 & \qt{23.71\textbf{m}}
        &  3.12 & \qt{41.61\textbf{s}}   & 26.40 & \qt{44.96\textbf{m}} \\
    \rowcolor{blue!10}\cellcolor{white}
      & \aname & PTQ
        & \textbf{78.84} & \qt{0.2\textbf{s}}   & \textbf{76.30} & \qt{0.53\textbf{s}}
        & 75.96 & \qt{0.2\textbf{s}}   & 75.76 & \qt{0.57\textbf{s}}
        & \textbf{93.05} & \qt{1.87\textbf{\textcolor{red}{s}}}
        & \textbf{89.88} & \qt{35.28\textbf{\textcolor{red}{s}}}
        & \textbf{39.03} & \qt{1.16\textbf{\textcolor{red}{s}}}
        & 61.83          & \qt{34.90\textbf{\textcolor{red}{s}}} \\
    \bottomrule
    \multicolumn{19}{r}{$*$Q.Time: Quantization Time,\; SGQ: SGQuant,\; DQ: Degree-Quant,\; \aname: Proposed Method}
  \end{tabular}
  }
\end{table*}

\begin{table}[t]
\centering
\caption{Quantized accuracy and time on GNN architectures with learnable edge weights}
\resizebox{.9\columnwidth}{!}{
\setlength{\tabcolsep}{3pt}  %
\begin{tabular}{l l 
                cc cc     %
                cc        %
                }
\toprule
      \multirow{3}{*}{Method}
      & \multirow{3}{*}{Type}
&\multicolumn{4}{c}{Cora} & \multicolumn{2}{c}{MAG240M} \\
\cmidrule(lr){3-6}\cmidrule(lr){7-8}
& & \multicolumn{2}{c}{INT8} & \multicolumn{2}{c}{INT4} & \multicolumn{2}{c}{INT8} \\
\cmidrule(lr){3-4}\cmidrule(lr){5-6}\cmidrule(lr){7-8}
&
  & Acc. & Q.Time 
  & Acc. & Q.Time
  & Acc. & Q.Time \\
\midrule

\rowcolor{gray!10}
FP32 & --- 
  & 80.36 & -- 
  & 80.36 & -- 
  & 69.66 & -- \\

\midrule
SGQ & QAT
    & 80.30 &\qt{7.7\textbf{s}} 
    & 77.92& \qt{6.8\textbf{s}}
    &46.76 & \qt{5.50\,\textbf{days}} \\
DQ & QAT  
  & 78.66 & \qt{14.6\textbf{s}}
  & 77.90 & \qt{14.5\textbf{s}}
  & N/A    & \qt{37.5\,\textbf{days}} \\

$A^{2}$Q & QAT  
  & 75.29 & \qt{6.3\textbf{s}}
  & 45.64 & \qt{6.3\textbf{s}}
  & 57.97 & \qt{2.46\,\textbf{days}} \\

DRA & PTQ  
  & 80.20 & \qt{3.6\textbf{s}}
  & 74.35 & \qt{3.3\textbf{s}}
  & 66.13 & \qt{2.06\,\textbf{days}} \\

\rowcolor{blue!10}
TopGQ & PTQ  
  & \textbf{80.63} & \qt{0.2 \textbf{s}}
  & \textbf{78.56} & \qt{0.2\textbf{s}}
    & \textbf{69.14} & \qt{58.8\,\textbf{\textcolor{red}{minutes}}} \\

\bottomrule
\end{tabular}}
\label{tab:gat}
\end{table}

\Cref{fig:overall} illustrates how we apply \Iname during quantization. %
In the calibration phase (\cref{fig:overall}a), we first compute $\Imath(v)$ for each node $v$, as defined in \cref{sec:dain}. 
For each value, we calculate node-wise quantization parameters $(s_v, z_v)$. %
If multiple nodes share the same \Iname, %
we aggregate the statistics by taking the global maximum and %
minimum (\cref{fig:overall}b), ensuring the quantization parameters cover the full %
range.
This gives a pair of quantization parameters for each unique \Iname. %
Finally, in inference, we only need to compute the $\Imath(v)$ for each unseen node $v$ and use it as a key to retrieve the appropriate quantization parameters (\cref{fig:overall}c).
For this, we retrieve the $k$-nearest TopPIN groups and interpolate among their parameters.
Such design leverages the finding that nodes with similar $\Imath(v)$ values exhibit similar activation distribution, as we theoretically demonstrated.

\section{Experimental Results}
\subsection{Experimental Settings}
We evaluate \aname on node-level and graph-level tasks against three QAT baselines: 
SGQuant~\citep{sgquant}, 
\dq~\citep{degreequant} (DQ), 
$A^2Q$~\citep{a2q}, 
and one recent PTQ baseline: DRA~\citep{dra}. 
For node classification, we use Cora, CiteSeer, Reddit, ogbn-products, and MAG240M.
For graph classification, we use IMDB-BINARY, and COLLAB, and report 10-fold cross-validation accuracy. %
We calibrate fully-trained GCN~\citep{GCN}, GraphSAGE (SAGE)~\citep{graphsage}, GIN~\citep{gin}, and GAT~\citep{gat} for 4-bit and 8-bit integer quantization, 
with a fixed bitwidth across all layers for fair comparison.
All datasets except MAG240M were evaluated in the inductive setting, better reflecting practical quantization use. 
MAG240M was evaluated in the original transductive setting, 
using R-GAT architecture from the ogb-lsc~\citep{hu2021ogblsc} challenge.
All experiments are conducted %
using %
A6000 GPU, RTX 4090 GPU,
and Intel(R) Xeon(R) Gold 6442Y CPU.
To evaluate the practical deployment potential on edge devices, we used NVIDIA Jetson AGX Orin.
We report both the accuracy and the quantization time. %
Bold indicates the best accuracy, and red indicates quantization times with unit changes (e.g., h $\to$ m, m $\to$ s) by \aname.
\subsection{Evaluation Results of Node-level Tasks}

\Cref{tab:large_graph_node}
reports 
node classification results, %
where the graph size spans from small (Cora, Citeseer), middle (Reddit), to large (ogbn-products).
Across all sizes, TopGQ is the fastest while matching or exceeding the baseline accuracy:
Baselines take up to hours (16.35h, Reddit, GraphSAGE) for quantization whilst \aname takes less than a minute.
Notably in INT4, 
\aname achieves the best accuracy, with up to a 28.87\%p gain over the strongest baseline for GCN.

We also evaluate on GAT, a representative architecture for using dynamic edge weights, shown in \Cref{tab:gat}. 
\aname %
preserves the accuracy of the original model, demonstrating generalization across different GNN structures. 
On Cora, \aname achieves the best accuracy while keeping quantization time under half a second.
Notably in 4-bit quantization, we observe at most 32.92\%p accuracy gain compared to the baselines.
We further emphasize the benefit of \aname by using a hyper-scale graph with 240 million nodes (MAG240M).
We do not report accuracy for DQ because its training time (37.5 days) is prohibitive given our computational budget.
Other methods take at least 2.06 days, up to 5.50 days to quantize a GNN on such a hyper-scale graph, while \aname cuts it down to 58.8 minutes, showing at least 50$\times$ speedup.
At the same time, \aname presents a negligible difference to the FP32 model, %
setting a new standard for practical quantization of large-scale GNNs.

\subsection{Evaluation Results of Graph-level Tasks}

\begin{table}[t]
\centering
\caption{Comparison of quantization accuracy and time for the graph-classification datasets}
\label{tab:graph_classification_results}
\setlength{\tabcolsep}{2.6pt}
\renewcommand{\arraystretch}{1.02}
\resizebox{\columnwidth}{!}{
\begin{tabular}{cccccccccc}
\toprule
   \multirow{3}{*}{Method}
& \multirow{3}{*}{Type}
  & \multicolumn{4}{c}{IMDB-BINARY}
  & \multicolumn{4}{c}{COLLAB} \\
\cmidrule(lr){3-6}\cmidrule(lr){7-10}
  & 
  & \multicolumn{2}{c}{GCN}
  & \multicolumn{2}{c}{GIN}
  & \multicolumn{2}{c}{GCN}
  & \multicolumn{2}{c}{GIN} \\
\cmidrule(lr){3-4}\cmidrule(lr){5-6}\cmidrule(lr){7-8}\cmidrule(lr){9-10}
  & 
  & Acc.\ & Q.Time
  & Acc.\ & Q.Time
  & Acc.\ & Q.Time
  & Acc.\ & Q.Time \\
\midrule
\rowcolor{gray!10}
\textbf{FP32}  & --
     & 79.58 & --   & 79.72 & --
     & 82.54 & --   & 82.31 & --   \\
\midrule
 \textbf{INT8} \\
    SGQ  &QAT  & 68.28 & \qt{5.88\textbf{m}}  & 68.26 & \qt{6.74\textbf{m}}
            & 80.96 & \qt{39.35\textbf{m}} &81.80 & \qt{40.26\textbf{m}} \\
   DQ  & QAT   & 77.32 & \qt{8.98\textbf{m}}  & 76.00 & \qt{9.08\textbf{m}}
            & 82.30 & \qt{2.39\textbf{h}}  & 81.62 & \qt{2.30\textbf{h}} \\
   $A^2Q$ & QAT & 75.12 & \qt{3.24\textbf{m}}  & 75.97 & \qt{3.78\textbf{m}}
            & 64.10 & \qt{14.75\textbf{m}} & 80.21 & \qt{14.50\textbf{m}} \\
   DRA  &PTQ  & 78.88 & \qt{2.24\textbf{m}}  & 78.52 & \qt{2.29\textbf{m}}
            & 82.08 & \qt{11.49\textbf{m}} & 82.18 & \qt{10.19\textbf{m}} \\
\rowcolor{blue!10}
   \aname & PTQ & \textbf{79.34} & \qt{2.18\textbf{\textcolor{red}{s}}}
            & \textbf{79.50} & \qt{2.05\textbf{\textcolor{red}{s}}}
            & \textbf{82.52} & \qt{13.86\textbf{\textcolor{red}{s}}}
            & \textbf{82.28} & \qt{11.71\textbf{\textcolor{red}{s}}} \\
\midrule
 \textbf{INT4} \\
    SGQ  &QAT  & 67.64 & \qt{5.89\textbf{m}}  & 63.72 & \qt{6.71\textbf{m}}
            & 78.14 & \qt{38.87\textbf{m}} & 72.06 & \qt{40.44\textbf{m}} \\
   DQ   & QAT  & 76.02 & \qt{9.03\textbf{m}}  & 75.98 & \qt{9.22\textbf{m}}
            & 73.24 & \qt{2.40\textbf{h}}  & \textbf{77.61} & \qt{2.31\textbf{h}} \\
 $A^2Q$ & QAT & 74.09 & \qt{3.13\textbf{m}}  & 75.62 & \qt{3.79\textbf{m}}
            & 69.32 & \qt{14.94\textbf{m}} & 74.78 & \qt{14.40\textbf{m}} \\
  DRA  & PTQ  & 74.32 & \qt{2.22\textbf{m}}  & 70.28 & \qt{2.30\textbf{m}}
            & 64.16 & \qt{11.45\textbf{m}} & 66.24 & \qt{10.18\textbf{m}} \\
\rowcolor{blue!10}
  \aname & PTQ& \textbf{76.71} & \qt{2.08\textbf{\textcolor{red}{s}}}
            & \textbf{76.00} & \qt{2.13\textbf{\textcolor{red}{s}}}
            & \textbf{81.75} & \qt{13.85\textbf{\textcolor{red}{s}}}
            & 77.39 & \qt{11.71\textbf{\textcolor{red}{s}}} \\
\bottomrule
\end{tabular}}
\vspace{-2mm}
\end{table}

\Cref{tab:graph_classification_results} presents the graph-level classification results on IMDB-BINARY and COLLAB. 
\aname significantly improves quantization speed while maintaining task performance. 
For instance, while Degree-Quant is the strongest baseline in GCN COLLAB, it takes $2.40$ hours for quantization.
However, \aname shows superior accuracy while cutting down the overhead to $13.85$ seconds.
While \aname takes the least time to quantize,
in many cases \aname also shows the best accuracy with minimal degradation compared to FP32.
We attribute this to \aname's explicit integration of GNN-aware design, 
leveraging \Iname to effectively capture local topology, %
while QAT baselines neglect these properties.
Overall, the experimental results %
demonstrate that \aname provides a robust balance between accuracy and quantization speed, making it well-suited for both small and large-scale GNN tasks. 

\subsection{Evaluation Results of Inference Latency}
\label{subsec:eval:inference}

\begin{table}[t]
  \centering
  \setlength{\tabcolsep}{6pt}
  \renewcommand{\arraystretch}{0.9}
  \caption{GCN inference time (sec) on GPU and edge device using ogbn-products, with mini-batching}
  \label{tab:kernel_all}
  \resizebox{.96\linewidth}{!}{
  \begin{tabular}{l l 
                  c c   %
                  c c } %
    \toprule
    \multirow{2}{*}{Method} & \multirow{2}{*}{Type} 
      & \multicolumn{2}{c}{\textbf{RTX4090}} 
      & \multicolumn{2}{c}{\textbf{Jetson AGX Orin}} \\
    \cmidrule(lr){3-4} \cmidrule(lr){5-6}
      & & Time & Speedup & Time & Speedup \\
    \midrule
    \rowcolor{gray!10} FP32 & -- 
      & 34.51 & -- 
      & 754.09 & -- \\
    SGQ   & QAT 
      & 20.53 & 1.68$\times$ 
      & 470.89 & 1.60$\times$ \\
    DQ   & QAT 
      & 20.37 & 1.69$\times$ 
      & 463.96 & 1.63$\times$ \\
    $A^2Q$         & QAT 
      & 27.74 & 1.24$\times$ 
      & 635.15 & 1.19$\times$ \\
    On-the-fly PTQ & PTQ 
      & 27.73 & 1.24$\times$ 
      & 689.82 & 1.09$\times$ \\
    \rowcolor{blue!10}\aname & PTQ 
      & 20.53 & 1.68$\times$ 
      & 473.25 & 1.59$\times$ \\
    \bottomrule
  \end{tabular}}
\end{table}

\Cref{tab:kernel_all} reports the inference latency of \aname and baselines with the minibatch setting of ogbn-products.
Measurements were conducted on both GPU and edge devices, reflecting practical scenarios for quantized GNN deployment.
A key observation is that $A^2Q$ and on-the-fly PTQ are expensive and slow. 
This is because both methods require row-wise scans per intermediate activations to derive quantization parameters.
This highlights the importance of storing quantization parameters and retrieving them via an efficient mapping function.
\aname leverages 
\Iname, %
where computation incurs negligible overhead, enabling efficient inference (\cref{sec:dain}).

\subsection{Analysis on \Iname and Ablation Study}
\label{sec:analysis}

\begin{table}[t]
  \centering
  \setlength{\tabcolsep}{5pt}
  \renewcommand{\arraystretch}{0.85}
    \centering
    \caption{Accuracy and computation time comparison
      for various indexing strategies 
      on IMDB-BINARY}
    \label{tab:centrality_comparison}
    \resizebox{.88\linewidth}{!}{
      \begin{tabular}{cccccc}
        \toprule
        Bit   & Node Index    & GCN    & SAGE & GIN     & Time    \\ 
        \midrule
        \multirow{5}{*}{INT4} 
              & Naive PTQ     & 60.14  & 74.76 & 56.50    & --   \\
              & Betweenness   & 50.00  & 50.00 & 50.00   & 1.85s   \\ 
              & Closeness     & 72.90  & 75.36 & 67.78  & 1.48s   \\ 
              & Katz          & 69.34  & 74.26 & 72.58   & 20.04s  \\ 
        \cmidrule(lr){2-6}
              & \Iname        & \bfseries 76.71  & \bfseries  75.72 & \bfseries 76.00  & \bfseries 0.00059s  \\ 
        \bottomrule
      \end{tabular}
    }
  \end{table}

We assess the effectiveness of \Iname by comparing it against a naive PTQ strategy as well as commonly used centrality measures, including betweenness, closeness, and Katz centrality (\Cref{tab:centrality_comparison}). 
We report both accuracy and the total computation time required to perform indexing for all nodes.
The naive PTQ approach, which %
utilize a single global quantization parameter, shows significant accuracy degradation due to high variance in node magnitudes. 
The %
centrality measures may mitigate the %
degradation, compared to naive PTQ. 
However, they require costly %
graph traversal per node, making them impractical for %
inference.
In contrast, \Iname only depends on 1-hop neighborhood information, thereby significantly reducing the computational overhead. 
Despite its lightweight design, \Iname %
outperforms other %
baselines, highlighting its practicality and effectiveness as an indexing strategy for GNN quantization.

\begin{table}[t]
  \centering
    \centering
    \caption{Ablation Study of \aname}
    \label{tab:ablation}
    \resizebox{.8\linewidth}{!}
    {
      \begin{tabular}{l l c c c c}
        \toprule
        \multirow{2}{*}{Bit} 
          & \multirow{2}{*}{Method} 
          & \multicolumn{2}{c}{Reddit} 
          & \multicolumn{2}{c}{ogbn-products} \\
        \cmidrule(lr){3-4} \cmidrule(lr){5-6}
          &  & GCN & SAGE & GCN & SAGE \\
        \midrule
        \multirow{3}{*}{INT4} 
        &Naive PTQ        & 3.79	&	2.97	&1.33		&	24.74	\\
        &Only \Iname      & 93.05	&	85.83	&	1.43	&	52.45	\\
        \cmidrule(lr){2-6}
        &TopGQ             & \textbf{93.05}	&	 \textbf{89.88}	&	 \textbf{39.03}	&	 \textbf{63.18}	\\
        \bottomrule
      \end{tabular}
    }
\end{table}

\cref{tab:ablation} shows the ablation study, 
where each row corresponds to the incremental addition of \Iname and selective \absorb to the naive PTQ baseline, ultimately forming \aname. 
While naive PTQ fails to exploit quantization bins, %
\Iname mitigates this limitation with topology. %
However, as graph size increases (e.g., ogbn-products, GCN), TopPIN alone proves insufficient. 
By adding \absorb, the node-wise quantization effects are preserved across layers, leading to additional accuracy recovery.

\section{Related Work}

\label{sec:related_work}
\textbf{GNN quantization}
efficiently reduces extensive memory and computational costs of GNNs~\citep{GCN, gat, gin, graphsage}.
\dq~\citep{degreequant} is the first work to quantize GNN using QAT,
excluding high-degree node activations in calibration for robust quantization parameters
and compressing later at inference. 
SGQuant~\citep{sgquant} and $A^2Q$~\citep{a2q} are also QAT methods, but they
differ in that they allow mixed-precision
to assign a higher bitwidth to high-magnitude features. 
The quantization parameters are optimized with gradients in 
DRA~\citep{dra} to reconstruct the FP32 distributions.
Thus, they require significant and redundant quantization overheads, whereas \aname allows orders of magnitude shorter quantization time. %

\textbf{Graph topology in GNNs}
is often integrated during training to help the model effectively learn the structural information~\citep{graphunet, featuregcn, hu2022hybrid, moleculenet, you2021identity, brasoveanu2023extending}. 
For example, \citep{graphunet} 
uses degree centrality to find central nodes for effective representation learning. 
Also, \citep{featuregcn} uses betweenness centrality to assign weights to each node at aggregation.
There are prior attempts to use topology for GNN binarization \citep{bnn1, bnn2}. 
However, these methods do not incorporate topology in relation to node feature patterns for GNN quantization.

\section{Conclusion}
We introduce \aname, a topology-aware PTQ framework for GNNs, eliminating retraining costs while preserving task accuracy. 
By leveraging a novel node index (\Iname) and \absorb, 
\aname handles unseen node features of differing magnitudes.
The node-level strategies enable fast and precise quantization while preserving the computational benefits of integer operations.
Experiments across various GNN architectures and datasets 
show that \aname achieves QAT-level accuracy, while reducing quantization time by an order of magnitude compared to prior works. 
\begin{acks}
This work was supported by the National Research Foundation of Korea (NRF) grant funded by the Korea government (MSIT) (RS-2026-25495605), %
the Institute of Information \& communications Technology Planning \& Evaluation (IITP) (RS-2024-00395134, %
RS-2024-00347394, %
IITP-2025-RS-2023-00256081),  %
in part by Samsung Electronics (IO230407-05813-01),
and the Korea Basic Science Institute (National research Facilities and Equipment Center) grant funded by the Ministry of Science and ICT (No. RS-2025-00564840).
Jinho Lee is the corresponding author.
\end{acks}

\newpage
\bibliographystyle{ACM-Reference-Format}
\bibliography{sample-base}

\end{document}